\documentclass[sigconf,nonacm]{acmart}

\AtBeginDocument{%
  }

\usepackage{listings}
\usepackage{xcolor}
\usepackage{enumitem}
\usepackage{subcaption}
\usepackage{tikz}
\usetikzlibrary{positioning, arrows.meta, fit, backgrounds, shapes.geometric}

\lstdefinelanguage{JavaScript}{
  keywords={function, return, var, let, const, if, else, for, while, do, switch, case, break, continue, new, this, typeof, instanceof, in, of, class, extends, import, export, default, from, async, await, try, catch, throw, finally, yield},
  sensitive=true,
  comment=[l]{//},
  morecomment=[s]{/*}{*/},
  morestring=[b]",
  morestring=[b]'
}

\begin{document}

\title{Substrate-Portable Execution for Production LLM Workflows (Industry Track)}

\author{Atul Kulkarni}
\affiliation{%
  \institution{Amazon}
  \city{Seattle}
  \state{WA}
  \country{USA}
}
\email{atukulka@amazon.com}

\author{Tarun Gopinath}
\affiliation{%
  \institution{Amazon}
  \city{Seattle}
  \state{WA}
  \country{USA}
}
\email{tarungop@amazon.com}

\author{Vijay Rajakumar}
\affiliation{%
  \institution{Amazon}
  \city{Seattle}
  \state{WA}
  \country{USA}
}
\email{vijar@amazon.com}

\author{Shrikar Katti}
\affiliation{%
  \institution{Amazon}
  \city{Seattle}
  \state{WA}
  \country{USA}
}
\email{shrikar@amazon.com}

\author{Parthasarathy Govindarajen}
\affiliation{%
  \institution{Amazon}
  \city{Seattle}
  \state{WA}
  \country{USA}
}
\email{partgovi@amazon.com}

\begin{abstract}
Production LLM agents often run the same logic---including tool-calling loops, retrieval chains, and compositional workflows---in several operational modes. However, execution semantics are usually coupled to one runtime. We encountered this portability problem in Rufus, a conversational AI assistant that integrates a large catalog of tool endpoints and serves millions of Amazon customers~\cite{rufus2025blog}. Systems such as Rufus use three execution modes: real-time serving for customer traffic, asynchronous execution for background tasks, and batch execution for high-volume offline workloads such as evaluation and content pre-generation. Each mode has different service-level objectives (SLOs) and typically uses a separate runtime. Reusing orchestration designed for streaming forces asynchronous and batch workloads into a blocking, synchronous execution model. It also prevents these workloads from using batch inference APIs, which offer a 50\% discount at published prices~\cite{bedrockpricing}.

We present a binding-adaptive agent execution platform that separates workflow definition from execution substrate. Developers express a workflow once as a typed dataflow graph. The platform compiles that graph to in-process streaming for real-time serving, durable orchestration with AWS SWF~\cite{swf2012} for asynchronous execution, or distributed stream processing with Apache Flink~\cite{flink2015} for batch inference. No workflow code changes are required. The central abstraction represents LLM inference as a suspendable graph node whose semantics depend on the selected substrate: streaming delivery online, durable retry in asynchronous execution, and batched submission offline.

We validated dozens of production agent configurations across five orchestration patterns: single-inference RAG, iterative ReAct, compositional PreAct, conditional routing, and multi-agent deep research. These patterns represent common synchronous orchestration structures in production LLM systems. Across all three bindings, we found no detectable difference in output quality. Batch execution also reduced per-query inference cost in line with published batch API pricing~\cite{bedrockpricing} and operated alongside the streaming path at production scale.
\end{abstract}



\maketitle

\section{Introduction}
\label{sec:intro}

Production agents must support workloads with sharply different requirements. The logic that streams tokens to a customer with sub-second latency may also evaluate hundreds of thousands of queries offline, pre-generate content for frequent requests, and run traffic-driven quality regressions. Existing agent frameworks generally do not adapt a single workflow definition to the distinct execution and optimization requirements of real-time, asynchronous, and batch workloads. We examine this problem in Rufus, a conversational AI assistant used by millions of Amazon customers~\cite{rufus2025blog}. Rufus integrates a large catalog of Model Context Protocol (MCP)~\cite{mcp2024} endpoints for functions such as product search, knowledge retrieval, and price tracking. Its agents use several orchestration patterns, including classifier-based routing, RAG-style retrieval, multi-step ReAct~\cite{react2023}, speculative PreAct~\cite{preact2025}, and multi-agent orchestration.

Production conversational AI systems commonly use three execution modes. The real-time path optimizes for time to first token and supports patterns such as tool-calling loops, RAG, and speculative execution. The asynchronous path handles background tasks that tolerate seconds or minutes of latency, including deep-research pipelines and offline enrichment. The batch path serves high-volume offline workloads such as prompt evaluation, quality regression testing, and content pre-generation. Each mode has distinct SLOs and typically runs on a different runtime: a streaming service, a durable workflow engine, or a stream-processing or job-scheduling system.

These modes share prompts, tools, and orchestration logic, but their runtimes are heterogeneous. When asynchronous and batch workloads reuse orchestration built for real-time serving, they also inherit its blocking execution model. This coupling prevents the use of batch inference APIs and places offline workloads on infrastructure provisioned for low-latency serving. Offline inference can represent a material share of total cost, while cloud providers price batch inference below real-time inference~\cite{bedrockpricing}. Accessing that batch path requires the architecture to separate orchestration from inference submission.

\paragraph{Our approach: execution-mode portability.} We define agent workflows once as typed dataflow graphs and compile them to several execution substrates. Declaring inference as a primitive, rather than hiding it inside a function call, lets the platform select a substrate-native mechanism: blocking execution with streaming delivery for real-time serving, asynchronous SWF task execution with durable orchestration, or queued batch submission for offline execution. We validated the platform across most agent configurations in Rufus's production registry and across all three execution modes. This paper makes three contributions. First, it identifies substrate-opaque inference as the structural barrier to adapting one workflow across real-time, asynchronous, and batch execution (\S\ref{sec:design}). Second, it presents a binding-adaptive architecture, including a typed dataflow DSL, a compilation model, and the rationale for first-class inference (\S\ref{sec:design}). Third, it reports on two unavoidable sources of divergence: time-sensitive tool outputs and substrate-specific infrastructure failures (\S\ref{sec:experience}).

\begin{table}[t]
\caption{Rufus inference workloads. All share the same agent logic; they differ in execution requirements. The ``Binding'' column shows the target substrate in the new platform.}
\label{tab:workloads}
\small
\begin{tabular}{lrll}
\toprule
\textbf{Workload} & \textbf{Infer/inv} & \textbf{Latency} & \textbf{Target Binding} \\
\midrule
Real-time serving & 1--5 & Sub-second TTFT & Local \\
Evaluation & 1--5 & Hours & Flink \\
Regression evaluation & 1--5 & Hours & Flink \\
Cache generation & 1--3 & Hours & Flink \\
Deep research & 5--12 & Minutes & SWF \\
\bottomrule
\end{tabular}
\end{table}

\section{System Design}
\label{sec:design}

\paragraph{Design challenge.} When asynchronous and batch workloads reuse real-time orchestration, they inherit its blocking, synchronous execution model. Replacing this model with asynchronous batch submission would require changing the shared orchestration loop needed for progressive token streaming. Independent invocations also cannot aggregate inference requests into a common batch. Per-workflow systems such as Temporal~\cite{temporal2023} and Step Functions~\cite{stepfunctions2016}, and agent frameworks such as LangGraph~\cite{langgraph2024} and Strands~\cite{strands2025}, treat each LLM call as an opaque blocking operation without substrate awareness. This coupling has broader operational costs: quality evaluation uses infrastructure optimized for time to first token rather than throughput, and features developed for one mode cannot be reused directly in another.

\subsection{Key Problems and Architecture}

The design addresses two key problems. First, workflow logic must be decoupled from substrate selection so that one definition can execute in different operational modes. Second, the binding compiler must control the inference strategy because it varies across substrates. If inference remains hidden inside an opaque function call, every substrate inherits the same blocking execution strategy.

\begin{figure*}[t]
\centering
\resizebox{\textwidth}{!}{%
\begin{tikzpicture}[
  node distance=0.5cm and 0.4cm,
  every node/.style={font=\small},
  layer/.style={draw, rounded corners, minimum width=12cm, minimum height=0.8cm, align=center},
  binding/.style={draw, rounded corners, minimum width=3.2cm, minimum height=1.4cm, align=center, font=\scriptsize},
  arr/.style={-{Stealth[length=2mm]}, thick},
]

\node[layer, fill=blue!8] (author) {Agent Logic (Java Combinator API \quad/\quad Script DSL --- hot-reloadable, same graph)};

\node[layer, fill=green!8, below=of author] (rasp) {Binding-Agnostic Workflow Graph (ASP): \texttt{map} · \texttt{infer} · \texttt{iterate} · \texttt{branch} · \texttt{compose}};

\node[layer, fill=orange!8, below=of rasp] (binder) {Binding Compiler (one-pass graph walk $\rightarrow$ substrate-native \texttt{BoundGraph})};

\node[binding, fill=yellow!10, below left=0.8cm and 2.2cm of binder] (local) {\textbf{Local Binding}\\\footnotesize Streaming token delivery\\\footnotesize Sub-second TTFT\\\footnotesize Customer-facing traffic};
\node[binding, fill=purple!10, below=0.8cm of binder] (swf) {\textbf{SWF Binding}\\\footnotesize Durable async execution\\\footnotesize Per-invocation traceability\\\footnotesize Deep research, reliability};
\node[binding, fill=red!10, below right=0.8cm and 2.2cm of binder] (flink) {\textbf{Flink Binding}\\\footnotesize Cross-invocation batching\\\footnotesize Horizontal scaling\\\footnotesize Eval, cache, analysis};

\node[layer, fill=gray!10, below=2.8cm of binder, minimum width=12cm] (bedrock) {Amazon Bedrock Inference (sync Converse API / batch inference API)};

\node[draw, rounded corners, fill=cyan!8, minimum width=2.0cm, minimum height=2.5cm, align=center, font=\scriptsize, right=0.6cm of rasp] (cp) {Control\\Plane\\[3pt]\tiny Agent Config\\Registry\\Lifecycle\\MCP Tools};

\draw[arr] (author) -- (rasp);
\draw[arr] (rasp) -- (binder);
\draw[arr] (binder) -- (local);
\draw[arr] (binder) -- (swf);
\draw[arr] (binder) -- (flink);
\draw[arr] (local) -- (local |- bedrock.north);
\draw[arr] (swf) -- (swf |- bedrock.north);
\draw[arr] (flink) -- (flink |- bedrock.north);
\draw[arr, dashed] (cp) -- (rasp);

\end{tikzpicture}%
}
\caption{Platform architecture. Agent logic (top) is authored via the typed Java API or the script DSL; both produce the same binding-agnostic workflow graph. The binding compiler translates the graph to substrate-native operations. Each binding targets a distinct operational profile: streaming for live traffic, durable async for traceable execution, batch for scalable offline operations. All bindings share the same Bedrock inference layer.}
\label{fig:architecture}
\end{figure*}
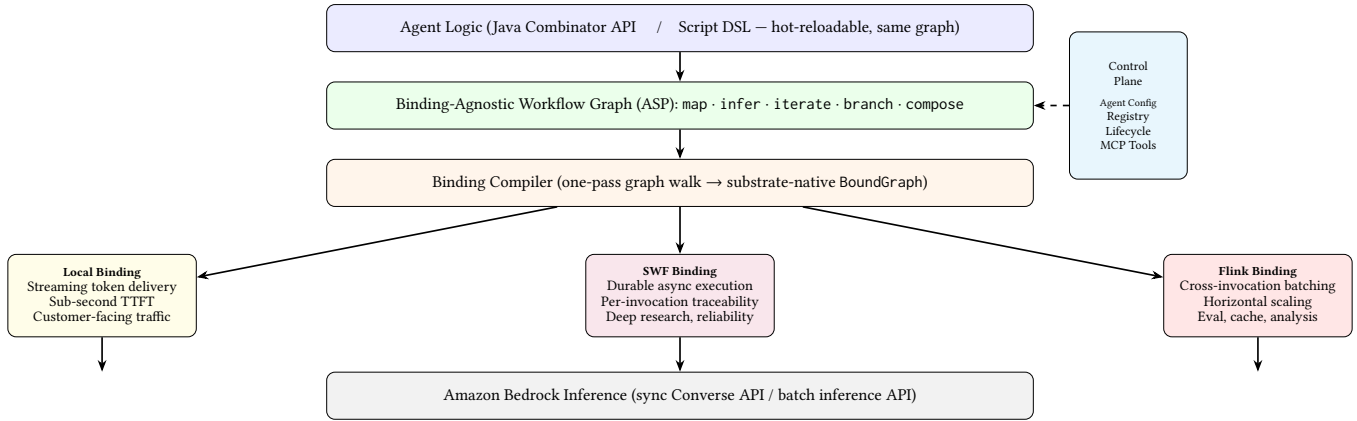

Figure~\ref{fig:architecture} shows the resulting architecture. Agent logic written in Java or JavaScript produces a binding-agnostic workflow graph. The binder compiles the graph into substrate-native operations for Local, SWF, or Flink execution. A shared control plane manages agent configuration, registration, and lifecycle across all bindings.

\subsection{Typed Dataflow DSL}

Agent workflows are typed dataflow graphs. The platform provides two equivalent authoring interfaces: a type-safe Java combinator API for production workflows and a hot-reloadable JavaScript DSL on GraalVM for exploration. These interfaces address different development needs: scripting supports rapid iteration, while the Java API provides compile-time checks for production code. We implemented both as frontends to the same combinator model and immutable graph rather than as separate execution paths. The binding system therefore treats their output identically, and the choice of authoring interface does not introduce different execution semantics.

\begin{lstlisting}[caption={A ReAct agent defined via the typed Java API. The JavaScript DSL produces the identical graph.},label=lst:workflow]
Workflow<AgentInput, AgentOutput> wf =
  Workflow.source("react-agent")
    .map(InitState())
    .iterate(ShouldContinue(), maxIter,
      step -> step
        .infer(ReasoningPrompt(), config)
        .map(ParseResponse())
        .map(ConditionalToolExec())
    )
    .map(BuildOutput());
\end{lstlisting}

Table~\ref{tab:combinators} summarizes the DSL's combinators, their semantics, and their typical uses.

\begin{table}[t]
\caption{Workflow combinators and their intended uses.}
\label{tab:combinators}
\small
\begin{tabular}{p{1.95cm}p{3.10cm}p{2.20cm}}
\toprule
\textbf{Operator} & \textbf{Semantics} & \textbf{Typical use} \\
\midrule
\texttt{map} & Applies a stateless transformation to node output. & Prompt assembly, parsing, tool dispatch \\
\texttt{infer} & Invokes an LLM using the strategy selected by the binding. & Model inference \\
\texttt{iterate} & Repeats a body until a condition or iteration limit is reached. & ReAct loops \\
\texttt{compose} & Delegates execution to another registered agent. & Multi-agent workflows, PreAct \\
\texttt{branch}, \texttt{parallelBranch} & Performs conditional or parallel fan-out followed by a typed merge. & Routing, specialist fan-out \\
\bottomrule
\end{tabular}
\end{table}

The small combinator set also expresses more complex patterns. Listing~\ref{lst:preact} shows speculative PreAct: a low-cost model plans tool use before delegating answer composition to a stronger ReAct agent.

\begin{lstlisting}[float=t,caption={PreAct composition in the JavaScript DSL. The same graph executes on every binding.},label=lst:preact,language=JavaScript]
workflow.source(Source("preact-compose"))
  .map(InitState())
  .infer(PlannerPrompt(),
    StaticInferenceConfig({modelId:"haiku45"}))
  .map(ParseResponse())
  .map(ToolExecution())
  .map(HandoffPrompt())
  .map(BuildOutput())
  .map(OutputAsContext())
  .compose(Agent("react-agent"));
\end{lstlisting}

This example exercises \texttt{infer}, \texttt{map}, and \texttt{compose}. Restricting the interface enables behavioral equivalence across bindings; arbitrary I/O or runtime graph mutation would make that guarantee infeasible.

\subsection{Why Inference is a First-Class Primitive}
\label{sec:inference-primitive}

Many agent frameworks represent an LLM call like any other function call. The framework sees an operation such as \texttt{invoke(fn)} and blocks until it returns, imposing one execution strategy on every workload.

Representing inference as a graph node, through \texttt{.infer(...)} rather than an opaque call inside \texttt{.map(...)}, enables binding-specific compilation. Inference has three relevant properties. First, it dominates workflow cost and latency, so its execution strategy largely determines the workflow's profile. Second, it is a stateless request to a remote service and can therefore be submitted asynchronously and correlated by identifier. Third, requests from independent invocations can be aggregated for batch submission. The Local binding uses streaming delivery, SWF uses asynchronous activity-task execution, and Flink batches requests across invocations (\S\ref{sec:bindings}). An opaque function call does not expose these properties to the substrate.

\subsection{Binding Compilation}
\label{sec:bindings}

A \emph{Binder} traverses the workflow DAG and translates each node into binding-specific operations. The result is a \texttt{BoundGraph}, an executable representation for the target platform. Each binding serves a distinct operational profile.

\noindent\emph{Local binding (real-time streaming).} Each node becomes a direct call on the invoking thread. Inference runs synchronously and streams tokens to the client. This binding prioritizes latency, time to first token, and progressive rendering.

\noindent\emph{SWF binding (deterministic, traceable asynchronous execution).} Each node becomes an SWF activity task. A decider schedules activity tasks and applies retry policies. SWF records every decision in an event history that can be inspected, replayed, and audited without additional instrumentation. This binding supports per-invocation traceability, durable state for long-running operations, and moderate-volume asynchronous features.

\noindent\emph{Flink binding (scalable batch execution).} Each node becomes a DataStream operator. Although the workflow DSL presents a synchronous programming model, blocking an operator would stall the Flink pipeline. The binder therefore stores suspended invocation state in Flink keyed state and offloads long-running operations through SQS. An inference service groups queued requests and submits them to the Bedrock batch API; responses are correlated by invocation identifier and resume the suspended state. The binder similarly offloads tool calls marked with \texttt{@Offloadable} to a separate worker pool. We chose this design over Flink's native asynchronous I/O so that application workers retain control over tool timeouts and retries, while the platform manages only suspension and resumption. This design enables cross-invocation batching, which per-workflow systems cannot provide, without changing workflow code or blocking operator threads. The binding prioritizes throughput and cost efficiency for evaluation and content-generation workloads.

\section{Evaluation}
\label{sec:eval}

We evaluate whether identical workflows produce statistically indistinguishable output quality across the Local, SWF, and Flink bindings. We then compare the cost and operational profiles of the three execution modes.

\subsection{Experimental Setup}

To isolate the platform from any specific production configuration, we use publicly available Anthropic Claude models on Amazon Bedrock: Opus, Sonnet, and Haiku. We evaluate five orchestration patterns that collectively exercise the DSL's combinators: (i)~\textbf{RAG}, a single inference over a large context without tool use (Opus); (ii)~\textbf{ReAct}, an iterative tool-calling loop with 2--5 iterations (Sonnet); (iii)~\textbf{Routing}, in which a classifier dispatches requests to specialist agents by intent (Haiku + Sonnet); (iv)~\textbf{PreAct}, in which a lower-cost planner selects tools speculatively and a stronger model composes the result (Haiku + Sonnet); and (v)~\textbf{Deep Research}, in which a planner creates tasks, specialists run in parallel, and a synthesizer composes the result (Sonnet, 5--12 inference calls). These patterns include single- and multi-model configurations as well as single- and multi-agent configurations.

The evaluation set contains more than 500 expert-authored and synthetic prompts per pattern and covers the query types targeted by the platform. The same benchmark is used to evaluate all production releases before deployment. We run all three bindings against the same Bedrock endpoints in one AWS region: Local on ECS Fargate, an SWF decider and activity workers on ECS, and Flink on Amazon Managed Flink with a horizontally scaled TaskManager cluster. We measure cross-binding equivalence using position-swapped pairwise preferences and a Bradley-Terry model (\S\ref{sec:equivalence}). We also measure per-query cost and end-to-end latency.

\subsection{Cross-Binding Quality Equivalence}
\label{sec:equivalence}

Each agent runs on the same dataset under Local streaming, SWF asynchronous execution, and Flink batch execution. We define quality equivalence as statistically indistinguishable outputs across these bindings.

\paragraph{Quality equivalence.} We follow Chatbot Arena's pairwise evaluation method~\cite{zheng2023judging,chiang2024chatbot} and control for position bias~\cite{shi2025positionbias,arenahard2024}. Each pair is judged in both orders. We declare a winner only if the same output wins in both orders, and call the resulting measure the \emph{decisive rate}. We fit a Bradley-Terry model~\cite{bradleyterry1952} to these preferences, which assigns each binding a strength $\beta$ on a log-odds scale such that the difference between two bindings' strengths gives the log-odds that one is preferred over the other. We fix $\beta_{\text{Stream}}=0$ as the reference, so a binding's $\hat{\beta}$ measures its preference relative to Stream: $\beta=0$ denotes equal preference and larger $|\beta|$ a stronger systematic preference. A 95\% bootstrap confidence interval over 1000 rounds that contains zero indicates no reliable preference. Table~\ref{tab:equivalence-bt} reports the results.

\begin{table*}[t]
\caption{Quality equivalence (Bradley-Terry). \textbf{Cross-bind decisive\%}: fraction of position-swapped comparisons where one binding wins in both orderings (maximum across the 3 binding pairs). $\hat{\beta}$: BT strength relative to Stream ($\beta_{\text{Stream}} \equiv 0$) with 95\% CI; CI containing zero = no detectable preference.}
\label{tab:equivalence-bt}
\small
\begin{tabular}{llrrr}
\toprule
\textbf{Pattern} & \textbf{Model(s)} & \textbf{Cross-bind} & $\hat{\beta}_{\text{SWF}}$ [CI] & $\hat{\beta}_{\text{Batch}}$ [CI] \\
 & & \textbf{decisive\%} & & \\
\midrule
RAG (single-infer) & Opus & 0.0\% & $\approx 0$ & $\approx 0$ \\
Routing (branch) & Haiku + Sonnet & 21.5\% & $+0.11\;[-0.26, +0.50]$ & $+0.02\;[-0.38, +0.39]$ \\
ReAct (tool-calling) & Sonnet & 12.9\% & $+0.14\;[-0.59, +0.93]$ & $-0.43\;[-1.42, +0.53]$ \\
PreAct (composition) & Haiku + Sonnet & 40.4\% & $+0.25\;[-0.05, +0.57]$ & $-0.13\;[-0.45, +0.21]$ \\
Deep Research & Sonnet & 12.1\% & $-0.30\;[-1.15, +0.49]$ & $+0.34\;[-0.64, +1.44]$ \\
\bottomrule
\end{tabular}
\end{table*}

\paragraph{Findings.} Every reported 95\% Bradley-Terry confidence interval includes zero. We therefore find no statistically detectable quality preference for any binding. ReAct and Deep Research produce fewer decisive comparisons, resulting in wider confidence intervals. Routing and PreAct produce more decisive comparisons and narrower intervals.

\paragraph{Binding overhead.} Table~\ref{tab:latency} reports mean end-to-end latency normalized to the Stream baseline. SWF adds one decision-task round trip per iteration and takes $1.1$--$3.4\times$ as long as Stream. Flink takes $33$--$152\times$ as long, primarily because of provider-side batch scheduling rather than binding overhead. Deep Research has the smallest Flink multiplier because all bindings execute specialist fan-out in parallel. Each fan-out stage completes when its slowest inference completes, rather than accumulating the latency of every specialist call, so latency is better amortized across the concurrent calls. ReAct also benefits from caching its static system prompt and tool definitions after the first iteration, reducing its batch multiplier relative to Routing and PreAct.

\begin{table}[t]
\caption{Mean end-to-end latency normalized to the Stream baseline ($1.0\times$) for each pattern. SWF overhead reflects per-iteration decision-task round-trips. Flink latency is dominated by batch job scheduling at the provider, not binding overhead.}
\label{tab:latency}
\small
\begin{tabular}{lrrr}
\toprule
\textbf{Pattern} & \textbf{Stream} & \textbf{SWF} & \textbf{Flink} \\
\midrule
RAG (single-infer) & $1.0\times$ & $1.5\times$ & $79\times$ \\
Routing (branch) & $1.0\times$ & $1.8\times$ & $152\times$ \\
ReAct (tool-calling) & $1.0\times$ & $3.4\times$ & $96\times$ \\
PreAct (composition) & $1.0\times$ & $1.8\times$ & $116\times$ \\
Deep Research & $1.0\times$ & $1.1\times$ & $33\times$ \\
\bottomrule
\end{tabular}
\end{table}

\paragraph{Implication.} The Flink batch binding can run quality evaluations at a cost reduction consistent with published batch API pricing~\cite{bedrockpricing}. In our experiments, it still detected operationally relevant changes such as model downgrades and prompt regressions. Because the selected binding did not mask these quality differences, batch evaluation can replace more expensive streaming evaluation for the tested agent architectures.

\subsection{Cost Structure}
\label{sec:infra}

Given the observed quality equivalence, binding selection depends on cost, latency requirements, and operational characteristics rather than output quality.

\paragraph{Token cost dominates.} For a representative Sonnet ReAct agent with three iterations and tens of thousands of input and output tokens per iteration, tokens account for approximately 100\% of per-query cost in every binding. Compute, orchestration, messaging, and other infrastructure collectively account for less than 0.5\%. Published batch inference pricing~\cite{bedrockpricing} is therefore the primary workflow-level cost lever. Stream and SWF use the same real-time API and have the same per-token cost. SWF instead provides persistent event history, fault-tolerant resumption, and durable coordination across long waits. Its per-step scheduling latency, however, makes it unsuitable for traffic sensitive to time to first token. Flink separates offline batch requests from customer-facing real-time requests at the platform boundary. At low utilization, fixed Managed Flink costs increase per-query infrastructure overhead by roughly an order of magnitude, although it remains a low single-digit percentage of total cost.

\section{Limits of Portability}
\label{sec:experience}

Although successful executions show behavioral equivalence (\S\ref{sec:equivalence}), validation across the three substrates identified limits to that guarantee. The platform provides \emph{functional} portability: for the same input, workflows produce statistically indistinguishable outputs across substrates. It does not make latency, failures, or other operational behavior identical across bindings.

\paragraph{Time-sensitive tool outputs.} Tools that return wall-clock-dependent data, such as current prices or availability windows, may produce different results when batch execution occurs hours later. These results can change the continuation predicate. This divergence is a property of the tool rather than the binding.

\paragraph{Substrate-specific failures.} Each substrate has distinct failures, so failure handling is not fully portable. In synchronous execution, a completed tool call returns directly to the agent. Asynchronous and batch execution instead depend on a correlated response from an external worker or batch job. A correlation timeout can terminate the workflow before the LLM can handle the error. Batch execution also introduces collective failures, including partial results, aggregate token limits, and longer completion windows. These risks are acceptable for offline workloads without strict latency SLOs.

\paragraph{Portability boundary.} During migration, the restricted combinator set required workflows to isolate side effects in tool calls and separate orchestration from execution concerns. Workflows that cannot be expressed in the DSL depend on execution context, such as the timing of streamed tokens for user-interface behavior. These dependencies mark the boundary between portable and binding-specific logic.

\section{Related Work}
\label{sec:related}

Table~\ref{tab:related} compares our work with agent frameworks, durable-execution middleware, and cross-substrate compilers. The comparison reflects differences in design emphasis rather than deficiencies, and the cited systems continue to evolve.

Apache Beam~\cite{beam2015,beamdocs} introduced a ``define once, execute anywhere'' model for data pipelines. Its runners provide equivalent backends under a common correctness model. In contrast, our bindings intentionally target substrates with different latency, cost, and failure semantics: streaming, durable asynchronous execution, and batch processing. Agent frameworks such as LangGraph~\cite{langgraph2024,langgraphdocs}, Strands~\cite{strands2025,strandsdocs}, and AutoGen~\cite{autogen2023,autogendocs} represent LLM calls as function invocations and emphasize API-level composability. We expose inference as a graph node so that a compiler can select its execution strategy for each substrate. DSPy~\cite{dspy2024} compiles \emph{what to say} rather than \emph{how to execute} and is complementary to this work.

Temporal~\cite{temporal2023,temporaldocs} and Azure Durable Functions~\cite{durablefunctions2021} focus on durable asynchronous execution~\cite{beldi2020,exoflow2023}. Cross-invocation batching and streaming token delivery are outside their primary design goals. Flink Stateful Functions~\cite{flinkstatefun2020,flinkstatefundocs} supports stateful serverless workflows, while our work operates at the agent-programming layer through a typed DSL and first-class inference. Inference-serving systems such as vLLM~\cite{vllm2023}, Orca~\cite{orca2022}, Sarathi~\cite{sarathi2024}, and FlexGen~\cite{flexgen2023}, as well as speculative decoding~\cite{speculativedecoding2023}, optimize GPU use below the workflow layer and are also complementary.

Evaluation frameworks such as HELM~\cite{helm2023} and AgentBench~\cite{agentbench2023} separate evaluation from production execution. Our platform instead reuses the production workflow definition to reduce behavioral drift. Cloud batch inference APIs, including Bedrock~\cite{bedrock2024} and OpenAI~\cite{openaibatch2024}, target single-turn inference and do not directly execute tool-calling or multi-turn workflows. Batch execution of these workflows therefore requires separating inference submission from orchestration, as our binding compiler does.

\begin{table}[t]
\caption{Capability comparison of design emphasis across related systems. \checkmark~= native, $\sim$~= partial or supported via extension, --~= not a primary design focus. Assessments reflect each system's publicly available documentation as of August 2026; these ecosystems evolve rapidly.}
\label{tab:related}
\small
\begin{tabular}{lccccc}
\toprule
\textbf{System} & \textbf{Str.} & \textbf{Batch} & \textbf{Durable} & \textbf{X-inv.} & \textbf{Port.} \\
\midrule
LangGraph~\cite{langgraph2024,langgraphdocs} & \checkmark & -- & $\sim$ & -- & -- \\
Strands/AutoGen~\cite{strands2025,autogen2023,strandsdocs,autogendocs} & \checkmark & -- & $\sim$ & -- & -- \\
Temporal~\cite{temporal2023,temporaldocs} & -- & -- & \checkmark & -- & $\sim$ \\
Beam~\cite{beam2015,beamdocs} & $\sim$ & \checkmark & -- & \checkmark & \checkmark \\
Flink SF~\cite{flinkstatefun2020,flinkstatefundocs} & $\sim$ & -- & \checkmark & $\sim$ & -- \\
\midrule
\textbf{This work} & \checkmark & \checkmark & \checkmark & \checkmark & \checkmark \\
\bottomrule
\end{tabular}
\end{table}

\section{Conclusion}
\label{sec:conclusion}

We presented a binding-adaptive execution platform based on more than two years of operating Rufus across heterogeneous real-time, asynchronous, and batch runtimes. The platform represents inference as a first-class operation in a typed dataflow DSL and compiles one workflow definition to each runtime. Tests across dozens of production agent configurations and five orchestration patterns found no detectable binding-specific difference in output quality. These results cover Amazon Bedrock and conversational shopping; providers or domains with different response characteristics may exhibit different cross-binding behavior. The batch binding also provides access to batch inference pricing and separates offline workloads from real-time serving. More broadly, execution-mode differences can be handled at the runtime layer when the workflow interface is restricted enough to preserve equivalence.

\bibliographystyle{ACM-Reference-Format}
\bibliography{references}

\end{document}